\documentclass[10pt,twocolumn,letterpaper]{article}

\usepackage[table,dvipsnames]{xcolor}
\usepackage{tikz}  

\usepackage{iccv}
\usepackage{times}
\usepackage{epsfig}
\usepackage{graphicx}
\usepackage{amsmath}
\usepackage{amssymb}

\usepackage{subfiles}
\usepackage{flushend}

\usepackage{multicol}
\usepackage{booktabs} 
\usepackage{etoolbox,siunitx} 

\robustify\bfseries
\robustify\itshape

\usepackage{pgfplots} 
\usepackage{pgfplotstable} 
\pgfplotsset{compat=newest}
\usetikzlibrary{plotmarks}
\usetikzlibrary{colorbrewer} 

\usepackage[hidelinks]{hyperref}

\iccvfinalcopy 

\def\iccvPaperID{xxx} 

\ificcvfinal\fi

\definecolor{olivegreen}{RGB}{0,170,0}
\definecolor{darkred}{RGB}{220,100,10}
\definecolor{tealblue}{RGB}{20,100,200}

\newcommand{\osvos}{Caelles2017}
\newcommand{\msk}{Perazzi2017}
\newcommand{\ofl}{Tsai2016}
\newcommand{\bvs}{NicolasMaerki2016}
\newcommand{\fcp}{Perazzi2015}
\newcommand{\jmp}{Fan2015}
\newcommand{\hvs}{Grundmann2010}
\newcommand{\sea}{Ramakanth2014}
\newcommand{\tsp}{Chang2013}

\newcommand{\afs}{Vijayanarasimhan2012}
\newcommand{\jfs}{ShankarNagaraja2015}

\newcommand{\ours}{SGV}

\newcommand{\nosvos}{OSVOS}
\newcommand{\nmsk}{MSK}
\newcommand{\nofl}{OFL}
\newcommand{\nbvs}{BVS}
\newcommand{\nfcp}{FCP}
\newcommand{\njmp}{JMP}
\newcommand{\nhvs}{HVS}
\newcommand{\nsea}{SEA}
\newcommand{\ntsp}{TSP}

\newcommand{\nafs}{AFS}
\newcommand{\njfs}{JFS}

\newcommand{\J}{\mathcal{J}}
\newcommand{\F}{\mathcal{F}}
\newcommand{\T}{\mathcal{T}}

\definecolor{rowblue}{RGB}{220,230,240}

\begin{document}

\title{Semantically-Guided Video Object Segmentation\vspace{-4mm}}
\author{Sergi Caelles$^*$, Yuhua Chen$^*$, Jordi Pont-Tuset, and Luc Van Gool\\
Computer Vision Lab, ETH Z\"urich\\
{\small\url{http://vision.ee.ethz.ch/~cvlsegmentation/}}
}

\maketitle

\begin{abstract}
This paper tackles the problem of semi-supervised video object segmentation, that is, segmenting an object in a sequence given its mask in the first frame.
One of the main challenges in this scenario is the change of appearance of the objects of interest. Their semantics, on the other hand, do not vary. 
This paper investigates how to take advantage of such invariance via the introduction of a semantic prior that guides the appearance model.
Specifically, given the segmentation mask of the first frame of a sequence, we estimate the semantics of the object of interest, and propagate that knowledge throughout the sequence to improve the results based on an appearance model.
We present Semantically-Guided Video Object Segmentation (\ours), which improves results over previous state of the art on two different datasets using a variety of evaluation metrics, while running in half a second per frame.

\end{abstract}

\renewcommand{\thefootnote}{\fnsymbol{footnote}}
\footnotetext[1]{First two authors contributed equally.}
\section{Introduction}

\begin{figure}
\setlength{\fboxsep}{0pt}
\centering
\hfill
\begin{minipage}{0.45\linewidth}
\centering
\fbox{\includegraphics[width=\linewidth]{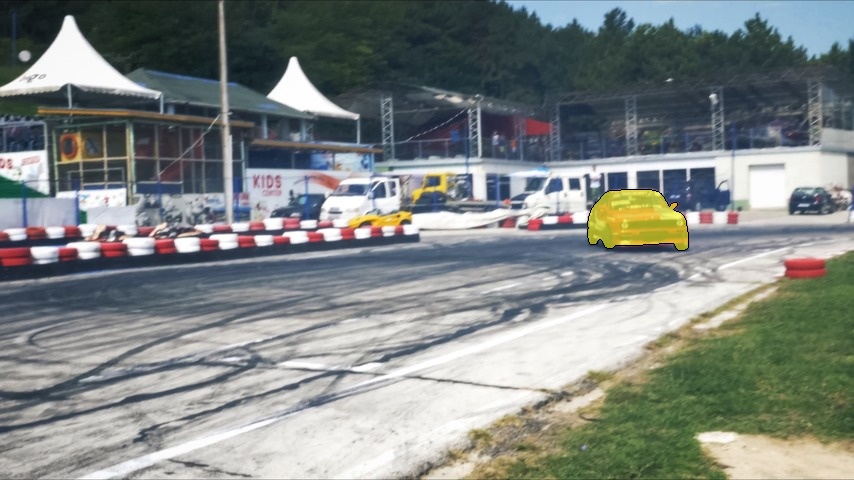}}\\
\scriptsize(a) First frame
\end{minipage}
\hfill\begin{minipage}{0.45\linewidth}
\centering
\fbox{\includegraphics[width=\linewidth]{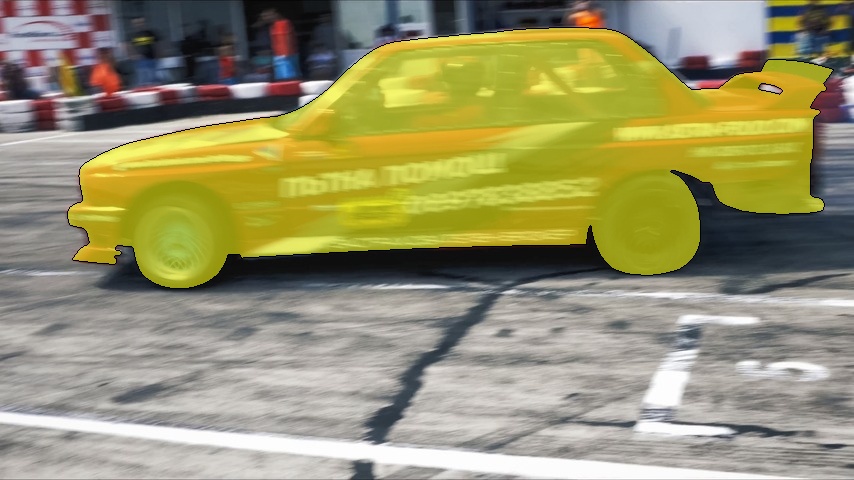}}\\
\scriptsize(b) Frame 34
\end{minipage}
\hfill\ \ %
\\[1mm]
\hfill
\begin{minipage}{0.45\linewidth}
\centering
\fbox{\includegraphics[width=\linewidth]{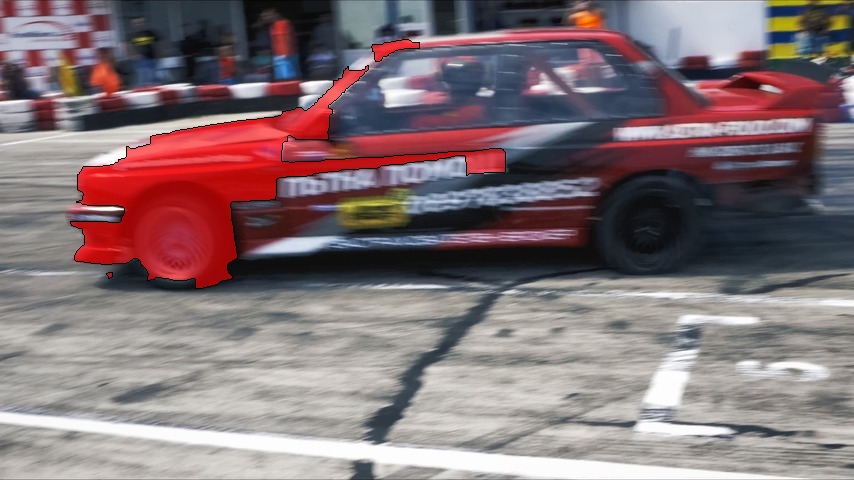}}\\
\scriptsize(c) Appearance
\end{minipage}
\hfill
\begin{minipage}{0.45\linewidth}
\centering
\fbox{\includegraphics[width=\linewidth]{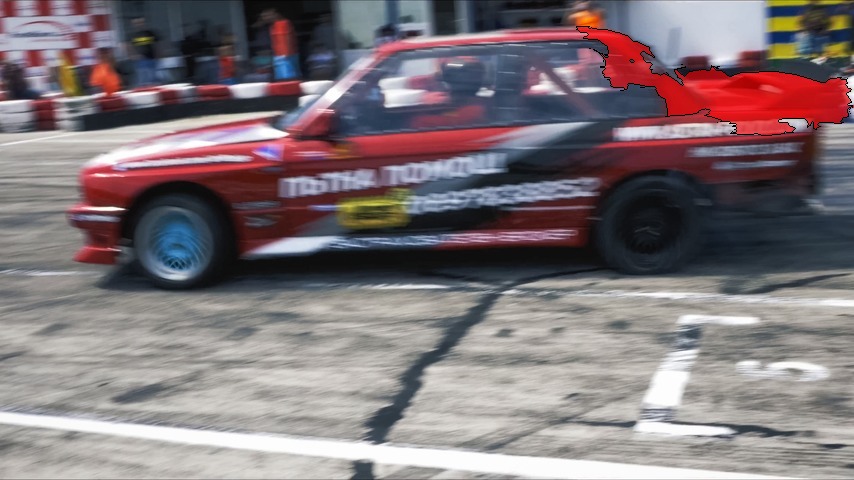}}\\
\scriptsize(d) Propagation
\end{minipage}
\hfill\ \ %
\\[1mm]
\hfill
\begin{minipage}{0.45\linewidth}
\centering
\fbox{\includegraphics[width=\linewidth]{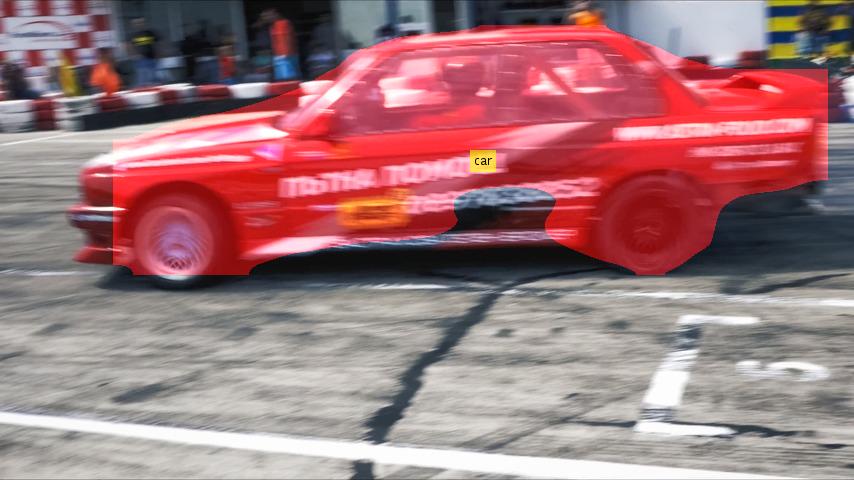}}\\
\scriptsize(e) Semantic prior
\end{minipage}
\hfill
\begin{minipage}{0.45\linewidth}
\centering
\fbox{\includegraphics[width=\linewidth]{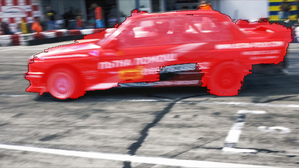}}\\
\scriptsize(f) Our result
\end{minipage}
\hfill\ \ %
\\[1mm]
\caption{\textbf{Motivation}: An object appearance can change significantly, and it can move too fast to follow. Its semantics, however, should not change throughout the sequence.}
\label{fig:example}
\end{figure}

This paper focuses on video object segmentation, which consists in outputting the mask of an object throughout a video sequence, given its segmentation in the first frame.
In the design of computer vision systems, human vision has always been the main source of motivation and inspiration; mimicking the way humans see is still elusive nowadays, although rapid progress is being done in the field.
So how do we humans solve video object segmentation?
If one is given the first image in Figure~\ref{fig:example}(a), with the mask of the object we want to track (in yellow), and is asked to give the segmentation in frame 34 (b), how would the mental process be like?
Most probably one thinks that \textit{one red racing car} has to be segmented, and then looks for it in the given frame.

Let us then review what the current state-of-the-art techniques do intuitively. 
First of all, \nosvos~\cite{Caelles2017} takes the first frame and builds a very accurate appearance model of the object using a Convolutional Neural Network (CNN).
It then classifies the pixels in the target frame according to the learnt appearance model and selects the ones that match the model.
Intuitively, in the previous example it would look for the \textit{specific red pixels it found in the first frame}. But what if the appearance has changed significantly because of light changes or dis-occlusions?
Figure~\ref{fig:example}(c) shows \nosvos' result in this case, where the dis-occlusion of the side of the car makes the mask to segment only the front.

Another approach is followed by MaskTrack~\cite{Perazzi2017}, which is trained on saliency datasets to segment the prominent object, given a rough mask estimate of it. 
Then, in video object segmentation, the mask of frame $n$ is taken as the estimate of the mask at frame $n+1$
and the algorithm refines the result to track the object. But what if the object moves very fast and so it changes position and shape dramatically between frames, or we do not have access to all frames of the sequence?
Figure~\ref{fig:example}(d) shows MaskTrack's result, where the fast movement causes the mask to drift out of the car.

This work is based on the idea of learning an appearance model of the object from the first frame (\textit{specific red pixels}) but, inspired by the human mental model, we also incorporate the semantic information that we are following \textit{one racing car}.
We therefore leverage an \textbf{appearance model} with a \textbf{semantic prior} and 
we refer to our method as \textit{Semantically-Guided Video Object Segmentation} (\ours).

So how to obtain the semantic prior?
We want to follow \textit{one} car, that is, a particular \textit{instance} of that category.
To do so, \ours{} makes use of an instance-level semantic segmentation algorithm, which goes beyond classifying pixels into categories and also groups them into different instances of that category.
Given the mask of the first frame, we select the top instances overlapping with the mask to decide \textit{what we track} (one instance of a car in the example).
Then we find the most overlapping instance mask or masks of the specific category with the appearance model result at frame~$n$ (Figure~\ref{fig:example}(e)).

To combine the appearance model result with the semantic prior, \ours{} uses a \textit{conditional classifier}, \ie, we build a background and foreground models that are combined given a foreground and background probability from the instance segmentation mask.
We implement the conditional classifier as a CNN layer that can be trained end to end.
The final result can be seen in Figure~\ref{fig:example}(f) where, interestingly, the appearance model has been able to refine the semantic prior in the front of the car, which was visible at the first frame, and has learnt to trust the semantic prior on the non-visible parts (side of the car).

\ours{} obtains state-of-the-art results both in DAVIS and Youtube-Objects using a variety of metrics. Quality barely decreases as the sequence advances thanks to the semantic prior, and the method runs in half a second per frame.
The contribution of this paper can be summarized as follows: i. We leverage a semantic prior in the domain of video object tracking, and we achieve state-of-the-art results based on different evaluations.
ii. To incorporate the semantic prior, we propose a novel conditional classifier layer, which is end-to-end trainable.

\section{Related Work}

\paragraph{Video Object Segmentation:}

As introduced previously, current literature in video object segmentation is dominated by two recent works.
First of all, Perazzi~\etal~\cite{\msk}, who use convolutional neural networks to learn to refine the mask of the object at frame $n$ to frame $n+1$. 
Previous to this work, enforcing temporal coherence was also a common practice to propagate the initial mask into the following frames.
Some works make use of superpixels~\cite{\tsp,\hvs}, patches~\cite{\sea,\jmp}, object proposals~\cite{\fcp}, or the bilateral filter~\cite{\bvs} to make the problem computationally feasible.
Optical flow is also used~\cite{\msk,\hvs,\sea} to match pixels from one frame to the next, which significantly penalizes execution speed.

Another approach was introduced by Caelles~\etal~\cite{\osvos}, in which a single appearance model from the object is learnt from the first frame, and then each frame is segmented independently.
This work follows this second line, since we believe it makes the algorithm robust to occlusions, to missing frames, and aligns better with the fact that humans can perfectly segment the object at frame $n$ directly from the first frame, without processing all the intermediate ones.
We differ from this work in that we augment the model with the semantic information of the object: what category and how many instances we are interested in.

\paragraph{Instance-Level Semantic Segmentation:}

Instance segmentation is a relatively new computer vision task which has gained increasing attention recently. In contrast to semantic segmentation or object detection, the goal of instance segmentation is to provide a segmentation mask for each individual instance. The task was first introduced in~\cite{hariharan2014simultaneous}, where they extract both region features and foreground features using R-CNN~\cite{girshick2014rich} framework and region proposals. Then, the features are concatenated and classified by SVM. Several works~\cite{dai2016instance,dai2015convolutional,zagoruyko2016multipath} following that path have been proposed in recent years. There also exist some approaches based on different techniques, such as: iteration~\cite{li2016iterative}, recurrent neural network~\cite{romera2016recurrent} or fully convolutional networks~\cite{li2016fully}.

\paragraph{Using Semantic Information to Aid Other Computer Vision Tasks:}
Semantic information is a very relevant cue for the human vision system, and some computer vision algorithms leverage it to aid various tasks. \cite{hane2013joint} improves reconstruction quality by jointly reasoning about class segmentation and 3D reconstruction.
Using a similar philosophy, \cite{liu2010single} estimates the depth of each pixel in a scene from a single monocular image guided by semantic segmentation, and improves the results significantly.
To the best of our knowledge, we are the first ones to apply semantic information to the task of object segmentation in videos.

\paragraph{Conditional Models:}
Conditional models prove to be a very powerful tool when the feature statistics are complex.
In this way, prior knowledge can be introduced by incorporating a dependency.
\cite{dantone2012real} builds a conditional random forest to estimate face landmarks whose
classifier is dependent on the pose of head.
Similarly, \cite{sun2012conditional} proposes to estimate human pose dependent on torso orientation, or human height, which can be a useful cue for the task of pose estimation.
The same also applies to boundary detection, \cite{uijlings2015situational} proposes to train a series of conditional boundary detectors, and the detectors are weighted differently during test based on the global context of the test image.
In this work, we argue that the feature distribution of foreground and background pixels are essentially different, and so a monolithic classifier for the whole image is bound to be suboptimal.
Thus, we utilize the conditional classifier to better model the different distributions.

\begin{figure*}
\includegraphics[width=\linewidth]{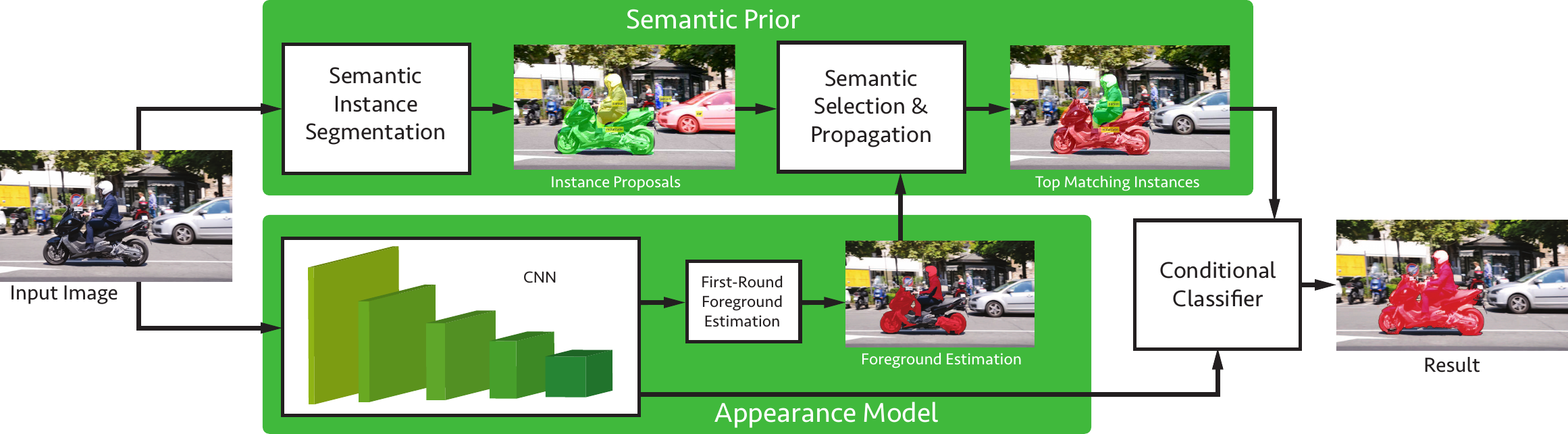}
\caption{\textbf{Network architecture overview}: Our network is composed of three major components: a base network as the feature extractor, and three classifiers built on top with shared features: a first-round foreground estimator to produce the semantic prior, and two conditional classifiers to model the appearance likelihood.}
\label{fig:overview_global}
\end{figure*}

\section{Semantically-Guided Video Object Segmentation (\ours)}

\begin{figure*}
\includegraphics[width=\linewidth]{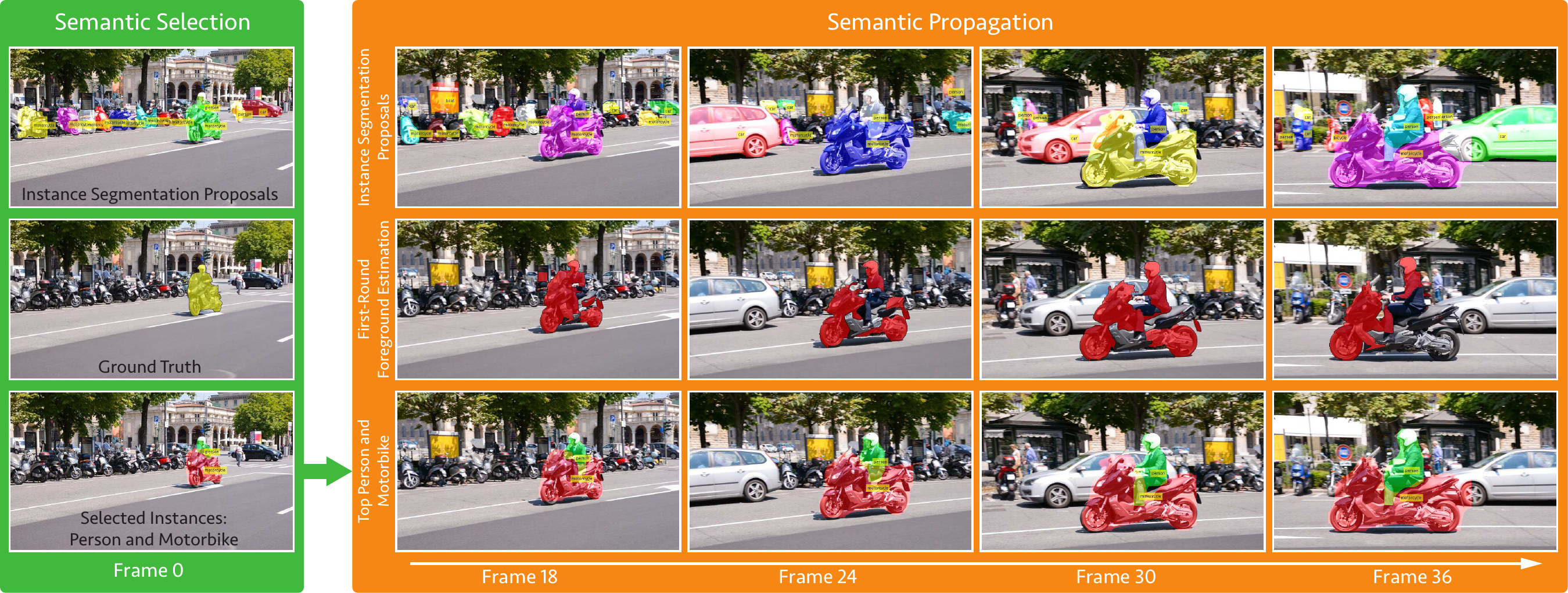}
\caption{\textbf{Semantic selection and propagation}: Illustrative example of the estimation of the semantics of the object from the first frame (semantic selection) and its propagation to the following frames (semantic propagation).}
\label{fig:overview_inst}
\end{figure*}

\subsection{Network Overview}

Figure~\ref{fig:overview_global} illustrates the structure and workflow of our proposed network. Following the common design of dense prediction networks~\cite{XiTu15,Maninis2016,Caelles2017}, we use VGG16~\cite{SiZi15} as our backbone network. The fully connected layers and the last pooling layer are removed to increase spatial feature resolution.
To aggregate multi-scale information from different layers, skip connections are added to extract \textit{hypercolumn} features. In more detail, we take the output feature maps from the second, third, fourth and fifth convolutional layer blocks before their corresponding pooling layers. Then, the feature maps are resized to the same size as the input image and we concatenate them to form the \textit{hypercolumn} features. Despite its simplicity, the architecture has shown to be highly effective in dense prediction tasks.

Sharing the common base network as the feature extractor, three pixel-wise classifiers are jointly learned. One classifier is for the first-round foreground estimation, which will be used in the semantic selection module to produce a semantic prior, together with the instance segmentation fed from an external instance segmentation system.
The other two classifiers act as appearance models: one responsible for the pixels that are part of the foreground while the other for background pixels. Finally, the two sets of predictions and the semantic prior are fused into the final prediction.

\subsection{Semantic Selection and Semantic Propagation}
We leverage a semantic instance segmentation algorithm as an input to estimate the semantics of the object to be segmented. Specifically, we choose MNC~\cite{dai2016instance} as our input instance segmentation algorithm due to its public implementation availability and good performance, but we are transparent to the algorithm used.

MNC~\cite{dai2016instance} is a multi-stage network that consists of three major components: shared convolutional layers, regional proposal network (RPN), and region-of-interest(ROI)-wise classifier. The model we use is pre-trained on Pascal, and outputs instance segmentation masks of 20 different classes.

The output of MNC is given as a set of masks which hopefully contain the objects of interest. We threshold the output to only choose from the ones with high confidence. Our objective is to find a subset of masks with consistent semantics throughout the video as our semantic prior. 

The process can be divided into two stages, namely \textit{semantic selection} and \textit{semantic propagation}.
Semantic selection happens in the first frame, where we select the masks that match the object according to the given ground-truth mask (please note that we are in a semi-supervised framework where the true mask of the first frame is given as input).
The number of masks and its categories are what we enforce to be consistent through the video.
Figure~\ref{fig:overview_inst} depicts an example of both steps.
Semantic selection, on the left in {\color{green!60!black}green}, finds that we are interested in a motorbike plus a person (bottom), by overlapping the ground truth (middle) to the instance segmentation proposals (top).

The semantic propagation stage (in {\color{orange}orange}) occurs at the following frames, where we propagate the semantic prior we estimated in the first frame to the following ones.
The instance segmentation masks (first row), are filtered using the first-round foreground estimation (middle row), and the top matching person and motorbike from the pool are selected (bottom row).

\subsection{Conditional Classifier}
Dense labeling using fully convolutional networks is commonly formulated as a per-pixel classification problem.
It can be therefore understood as a \textit{global} classifier sliding over the whole image,
and assigning either the foreground or background label to each pixel according to a \textit{monolithic} appearance model.
In this work, we want to incorporate the semantic prior to the final classification, which will be given as a mask of the most promising instance (or set of instances) in the current frame.

If semantic instance segmentation worked perfectly, we could directly select the best-matching instance to the appearance model, but in reality the results are far from perfect.
We can only, therefore, use the instance segmentation mask as a guide, or a guess, of what the limits of that instance are, but we still need to perform a refinement step.
Our proposed solution to incorporate this mask but still keep the per-pixel classification is to train two classifiers and weigh them according to the confidence we have in that pixel being part of the instance or not.
We argue that using a single set of parameters for the whole image is suboptimal.

Formally, for each pixel $i$, we estimate its probability of being a foreground pixel given the image:$p(i|I)$.
The probability can be decomposed into the sum of $k$ conditional probabilities weighted by the prior:
\[p(i|I)=\sum_{k=1}^{K} p(i|I,k)\,p(k|I).\]
In our experiments, we use $K=2$ and we build two conditional classifiers, one focusing on the \textit{foreground} pixels, and the other focusing on the background pixels.
The prior term $p(k|I)$ is estimated based on the instance segmentation output.
Specifically, a pixel relies more on the \textit{foreground} classifier if it is located within the instance segmentation mask; and more importance to the background classifier is given if it falls out of the instance segmentation mask. In our experiments, we apply a Gaussian filter to spatially smooth the selected masks as our semantic prior.

\begin{figure}
\includegraphics[width=\linewidth]{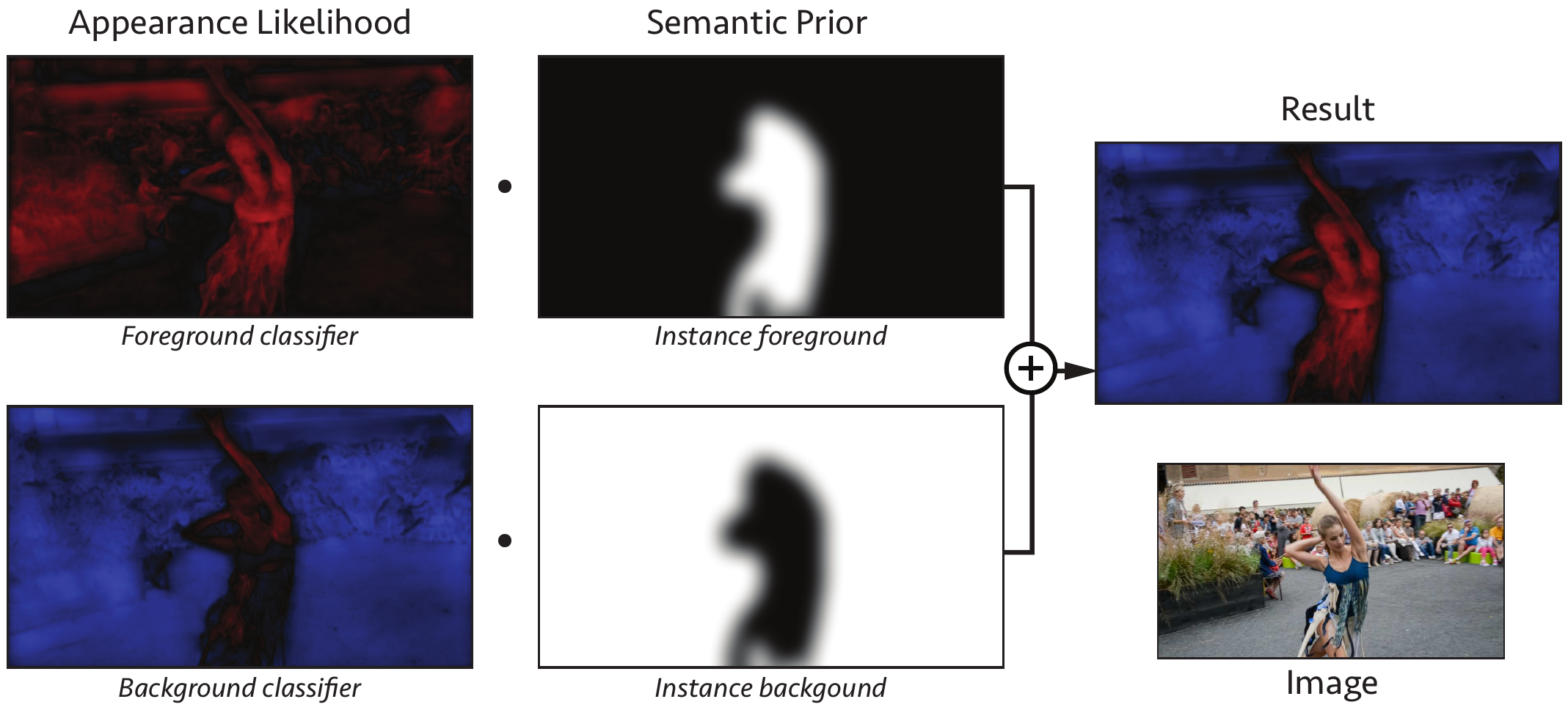}
\caption{\textbf{Forward pass of the conditional classifier layer}: {\color{red}Red} denotes foreground probability, and {\color{blue}blue} background probability. The output is the weighted sum of the two conditional classifier. }
\label{fig:classifier}
\end{figure}

The conditional classifier is implemented as a layer which can be easily integrated in the network in a end-to-end trainable manner. The layer take two prediction maps $f_1$ and $f_2$ and the weight map $w$ obtained using the semantic prior as inputs.
The inference process is illustrated in Figure~\ref{fig:classifier}, where each input element is multiplied by its corresponding weight from the weight map,
then summed with the corresponding element in the other map:
\[f_{out}(x,y) = w(x,y)\,f_{1}(x,y) + \big(1\!-\!w(x,y)\big)\,f_{2}(x,y).\]
Similarly, in the back-propagation step, the gradient from the top $g_{top}$ is propagated to the
two parts according to the weight map:
\begin{align*}
g_1(x,y)&= w(x,y)\,g_{top}(x,y)\\
g_2(x,y)&=\big(1-w(x,y)\big)\,g_{top}(x,y)
\end{align*}

\subsection{Training and Inference}
We use a two steps approach to train the network. First, the VGG part of the architecture is initialized using the weights pre-trained on Imagenet~\cite{Russakovsky2015}. We train the architecture to learn which are the pixels that are most probable to be foreground given a certain image. For this training step, we use the training set of the DAVIS dataset~\cite{Perazzi2016}. We also use side output supervision in the VGG part because it has proven to improve the convergence and the performance of the network~\cite{Maninis2016,XiTu15}.

In the second training step, we focus on learning the appearance model for the specific object that we want to segment in a video sequence. Therefore, this is done at test time initializing the CNN with the weights of the first step and fine-tuning it for a few iterations.

In order to produce the segmentation at every frame, we apply the fine-tuned network for the specific object of a video sequence to obtain the mask that corresponds to the object, with a single forward pass.

\section{Experimental Validation}

\paragraph*{Experimental Setup:}
We will mainly work on the DAVIS database~\cite{Perazzi2016}, using their proposed metrics:
region similarity (intersection over union $\J$),
contour accuracy ($\F$ measure), and temporal instability ($\T$).
The dataset contains 50 full-HD annotated video sequences, 30 in the training set and 20 in the validation set.
All our results will be trained on the former, evaluated on the latter.

We take the pre-computed evaluation results provided by the authors, which includes a large body of state-of-the-art techniques (\nofl~\cite{\ofl}, \nbvs~\cite{\bvs}, \nfcp~\cite{\fcp}, \njmp~\cite{\jmp}, \nhvs~\cite{\hvs}, \nsea~\cite{\sea}, and 
\ntsp~\cite{\tsp}).
We also take the two most recent works (\nosvos~\cite{\osvos} and \nmsk~\cite{\msk}) and evaluate their publicly-available pre-computed results.

For completeness, we also experiment on the Youtube-objects~\cite{Prest2012,Jain2014} dataset
against \nosvos~\cite{\osvos}, \nmsk~\cite{\msk}, \nofl~\cite{\ofl}, \nbvs~\cite{\bvs}, \nafs~\cite{\afs}, and \njfs~\cite{\jfs}.
We take the pre-computed evaluation results from previous work.

\begin{figure*}
\centering
\begin{minipage}{0.19\linewidth}
\centering
\includegraphics[width=\linewidth]{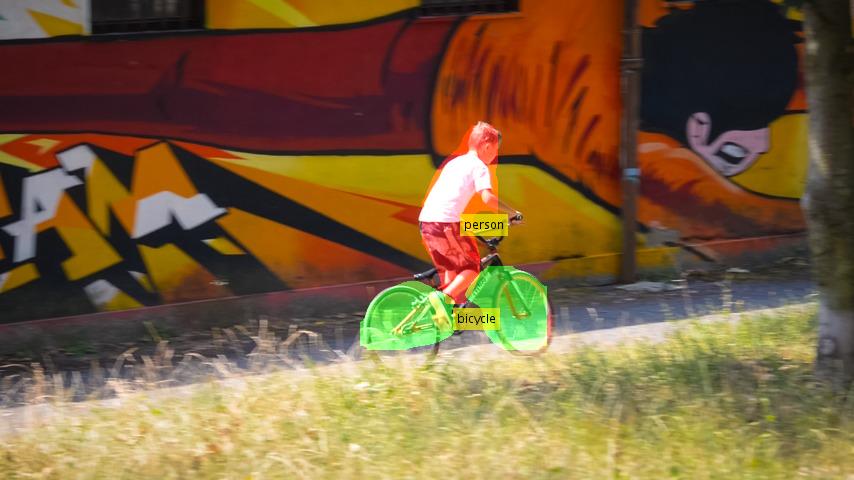}\\
\scriptsize(a) Person, bicycle
\end{minipage}
\hfill
\begin{minipage}{0.19\linewidth}
\centering
\includegraphics[width=\linewidth]{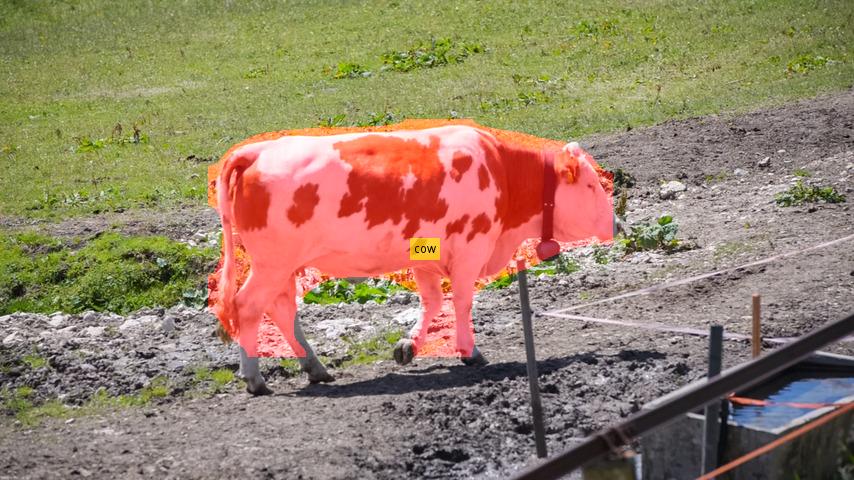}\\
\scriptsize(b) Cow
\end{minipage}
\hfill
\begin{minipage}{0.19\linewidth}
\centering
\includegraphics[width=\linewidth]{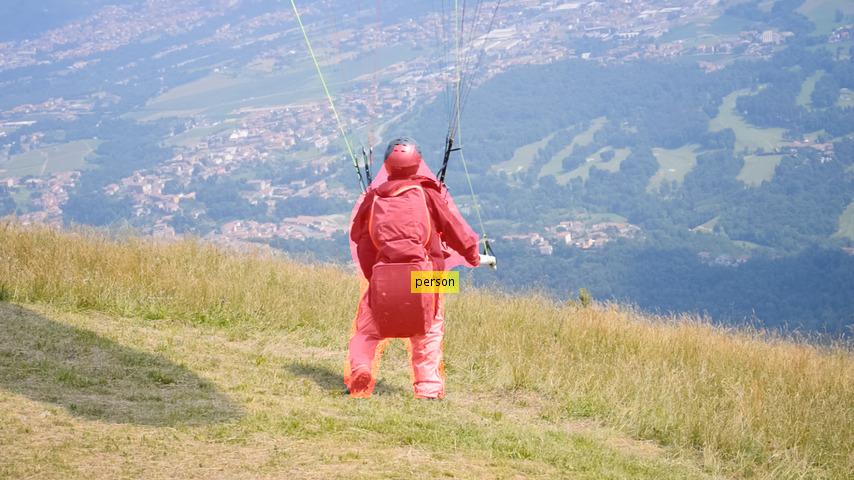}\\
\scriptsize(c) Person
\end{minipage}
\hfill
\begin{minipage}{0.19\linewidth}
\centering
\includegraphics[width=\linewidth]{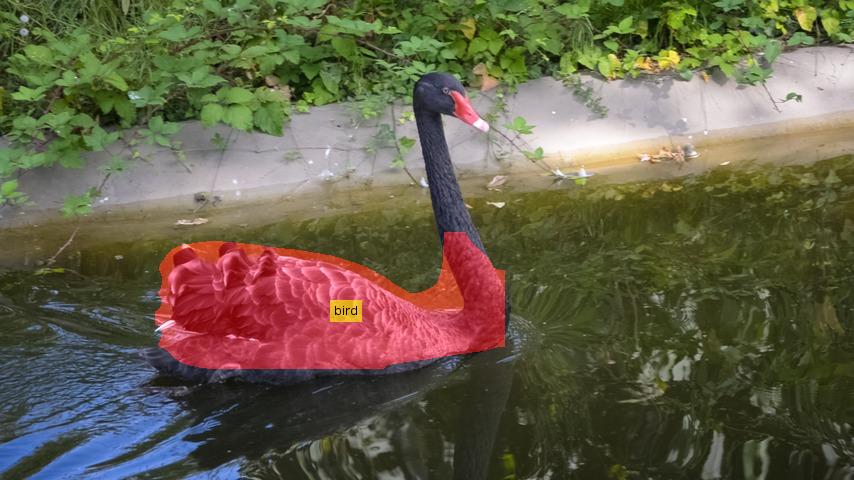}\\
\scriptsize(d) Bird
\end{minipage}
\hfill
\begin{minipage}{0.19\linewidth}
\centering
\includegraphics[width=\linewidth]{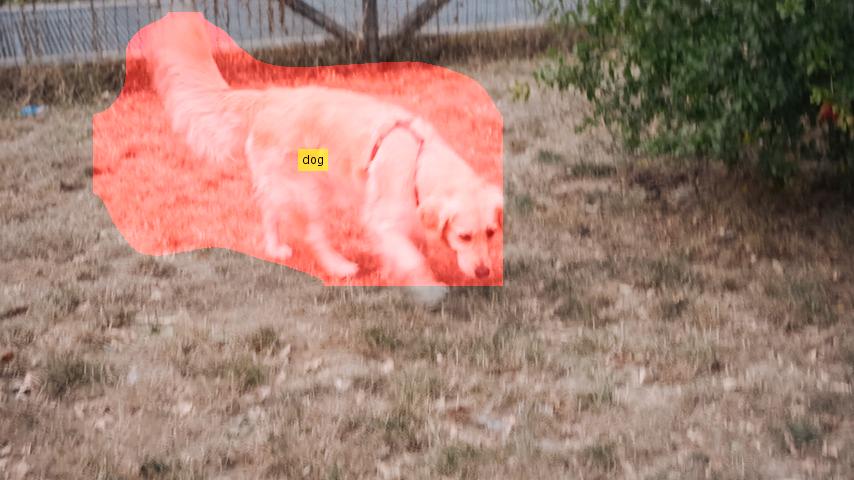}\\
\scriptsize(e) Dog
\end{minipage}
\\[1mm]
\begin{minipage}{0.19\linewidth}
\centering
\includegraphics[width=\linewidth]{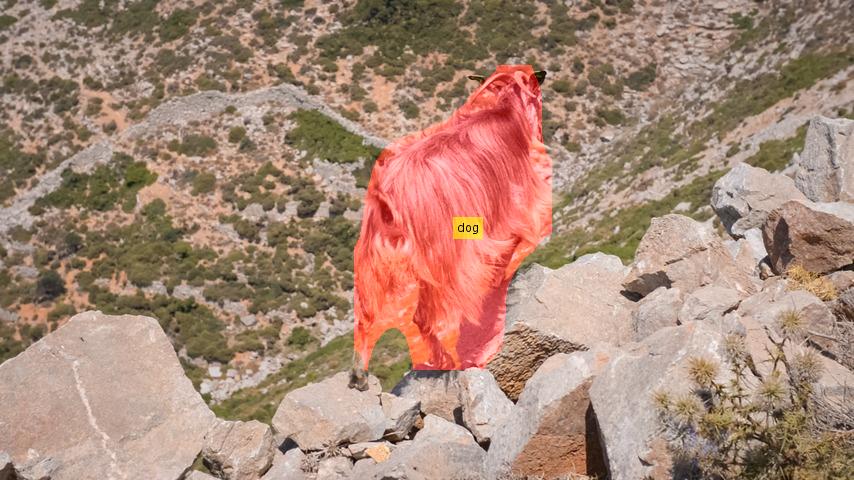}\\
\scriptsize(f) Dog
\end{minipage}
\hfill
\begin{minipage}{0.19\linewidth}
\centering
\includegraphics[width=\linewidth]{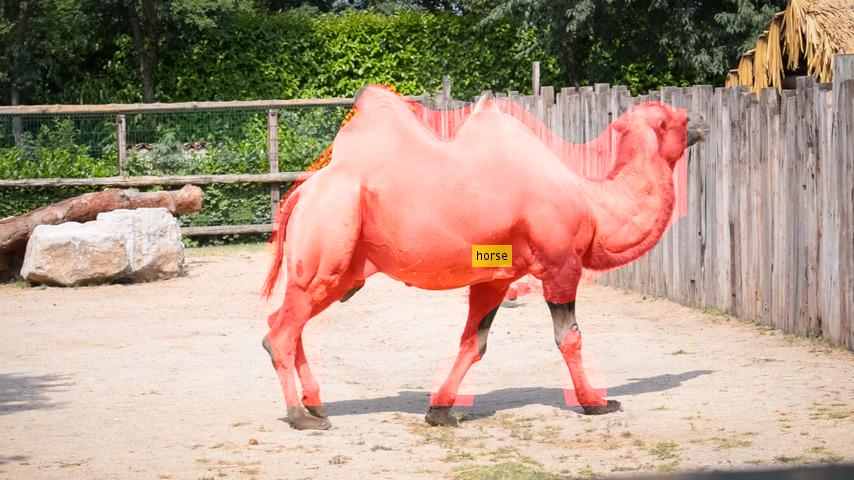}\\
\scriptsize(g) Horse
\end{minipage}
\hfill
\begin{minipage}{0.19\linewidth}
\centering
\includegraphics[width=\linewidth]{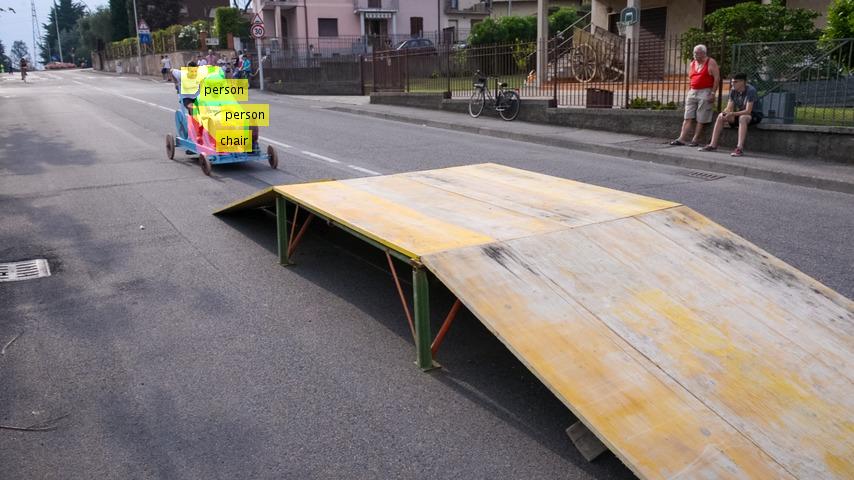}\\
\scriptsize(h) Person, person, chair
\end{minipage}
\hfill
\begin{minipage}{0.19\linewidth}
\centering
\includegraphics[width=\linewidth]{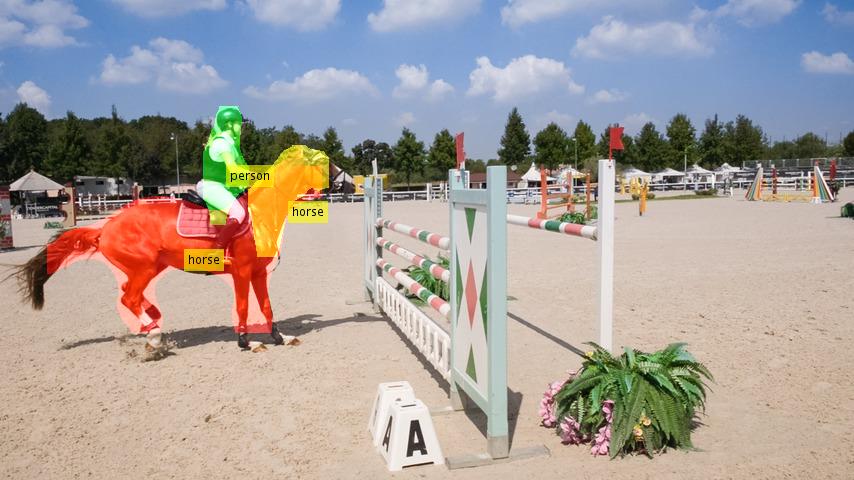}\\
\scriptsize(i) Horse, horse, person
\end{minipage}
\hfill
\begin{minipage}{0.19\linewidth}
\centering
\includegraphics[width=\linewidth]{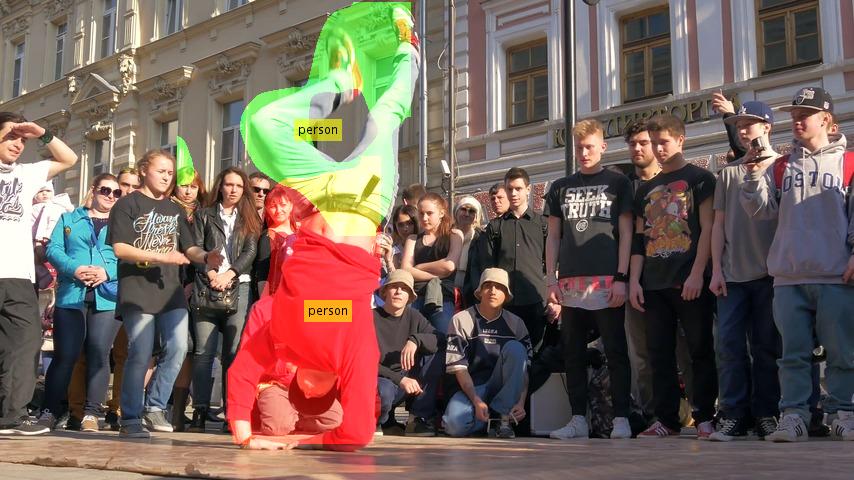}\\
\scriptsize(j) Person, person
\end{minipage}
\\[0.5mm]
\caption{\textbf{Semantic selection evaluation}: Semantic instances selected by our algorithm with its category overlaid.}
\label{fig:semantic_selection}
\end{figure*}

\paragraph*{Architectural Choices:}
The first design choice that we want to evaluate is the CNN architecture used to obtain the appearance model.
The simplest option is to use a feedforward network such as VGG-16 with skip connections (side outputs) as used in~\cite{XiTu15,Maninis2016,Caelles2017}.
Table~\ref{tab:architecture} (Skip) shows the results using this architecture.

\begin{table}
\setlength{\tabcolsep}{4pt} 
\center
\footnotesize
\rowcolors{3}{white}{rowblue}
\resizebox{0.75\linewidth}{!}{%
\sisetup{detect-all=true}
\begin{tabular}{llS[table-format=2.1]S[table-format=2.1]S[table-format=2.1]S[table-format=2.1]}
\toprule
						    &  &          & \multicolumn{3}{c}{Top Down} \\
\multicolumn{2}{c}{Measure} & \si{Skip} & \si{Naive} & \si{Side} & \si{Freeze} \\
\cmidrule(lr){1-2} \cmidrule(lr){3-3}\cmidrule(lr){4-6}
                                 & Mean $\mathcal{M} \uparrow$     &          77.4 &          64.1 &          68.0 &\bfseries 77.6 \\
\cellcolor{rowblue}$\mathcal{J}$ & Recall $\mathcal{O} \uparrow$   &          91.0 &          72.0 &          77.5 &\bfseries 91.3 \\
                                 & Decay $\mathcal{D} \downarrow$  &          17.4 &          23.3 &          23.5 &\bfseries 16.9 \\
\hline
                                 & Mean $\mathcal{M} \uparrow$     &\bfseries 78.1 &          63.1 &          66.9 &          77.5 \\
\cellcolor{white}$\mathcal{F}$ & Recall $\mathcal{O} \uparrow$   &\bfseries 92.0 &          70.3 &          75.9 &          91.3 \\
                                 & Decay $\mathcal{D} \downarrow$  &          19.4 &          23.5 &          24.1 &\bfseries 18.4 \\
\hline
\cellcolor{rowblue}$\mathcal{T}$ & Mean $\mathcal{M} \downarrow$   &\bfseries 33.6 &          54.7 &          51.3 &          35.6 \\
\bottomrule
\end{tabular}
}
\vspace{2mm}
\caption{\label{tab:architecture}
\textbf{Architecture choice evaluation for the appearance model}: Comparison between the feedforward+skip architecture and different training procedures when adding a top-down refinement branch.}
\vspace{0mm}
\end{table}

Motivated by SharpMask~\cite{Pinheiro2016}, we can also augment it with a top-down refinement modules, which we will evaluate in the following paragraphs.
In the first experiment we train the whole feedforward+top-down network naively, initializing the VGG with pre-trained Imagenet weights (Table~\ref{tab:architecture} Naive).
This does not reach the performance of Skip, probably due to the mix of the different paths of the gradients in the VGG layers from all the refinement modules.
To alleviate this effect, we incorporate two training improvements.

First, we add side supervision in the intermediate layers, as in~\cite{XiTu15}.
As can be seen in Table~\ref{tab:architecture} (Side), results improve, but they are still much lower than those of the feedforward+skip architecture.
To overcome this problem, we first train the feedforward layers alone with side supervision, and then we freeze the weights when learning the top-down refinement module.
With this, Table~\ref{tab:architecture} (Freeze) we are able to achieve slightly better in some metrics than with the feedforward+skip architecture.

As a conclusion, the minor improvements achieved using the top-down refinement architecture do not compensate the more complex and costly training, so we will select a feedforward network with skip connections as our architecture to obtain the appearance model.

\begin{table*}
\setlength{\tabcolsep}{6pt} 
\center
\footnotesize
\rowcolors{1}{white}{rowblue}
\resizebox{0.80\linewidth}{!}{%
\sisetup{detect-all=true}
\begin{tabular}{llS[table-format=2.1]S[table-format=2.1]S[table-format=2.1]S[table-format=2.1]S[table-format=2.1]S[table-format=2.1]S[table-format=2.1]S[table-format=2.1]S[table-format=2.1]S[table-format=2.1]}
\toprule
\multicolumn{2}{c}{Measure} & \si{Ours} & \si{\nosvos} & \si{\nmsk} & \si{\nofl} & \si{\nbvs} & \si{\nfcp} & \si{\njmp} & \si{\nhvs} & \si{\nsea} & \si{\ntsp} \\
\cmidrule(lr){1-2} \cmidrule(lr){3-12}
                                 & Mean $\mathcal{M} \uparrow$     &\bfseries 84.0 &          79.8 &          79.7 &          68.0 &          60.0 &          58.4 &          57.0 &          54.6 &          50.4 &          31.9 \\
\cellcolor{rowblue}$\mathcal{J}$ & Recall $\mathcal{O} \uparrow$   &\bfseries 96.1 &          93.6 &          93.1 &          75.6 &          66.9 &          71.5 &          62.6 &          61.4 &          53.1 &          30.0 \\
                                 & Decay $\mathcal{D} \downarrow$  &          4.9 &          14.9 &          8.9 &          26.4 &          28.9 &\bfseries -2.0 &          39.4 &          23.6 &          36.4 &          38.1 \\
\hline
                                 & Mean $\mathcal{M} \uparrow$     &\bfseries 83.0 &          80.6 &          75.4 &          63.4 &          58.8 &          49.2 &          53.1 &          52.9 &          48.0 &          29.7 \\
\cellcolor{white}$\mathcal{F}$ & Recall $\mathcal{O} \uparrow$   &\bfseries 93.8 &          92.6 &          87.1 &          70.4 &          67.9 &          49.5 &          54.2 &          61.0 &          46.3 &          23.0 \\
                                 & Decay $\mathcal{D} \downarrow$  &          7.7 &          15.0 &          9.0 &          27.2 &          21.3 &\bfseries -1.1 &          38.4 &          22.7 &          34.5 &          35.7 \\
\hline
\cellcolor{rowblue}$\mathcal{T}$ & Mean $\mathcal{M} \downarrow$   &          25.6 &          37.8 &          21.8 &          22.2 &          34.7 &          30.6 &          15.9 &          36.0 &\bfseries 15.4 &          41.7 \\
\bottomrule
\end{tabular}
}
\vspace{2mm}
\caption{\label{tab:soa}
\textbf{State-of-the-art comparison}: Comparison of \ours\ to the publicly-available semi-supervised results on DAVIS.
}
\vspace{0mm}
\end{table*}

\paragraph*{Semantic Selection and Propagation:}
The first step of our algorithm consists in selecting the semantic instance segmentations (Semantic Selection).
Since there is no ground truth available in DAVIS regarding semantics, we will evaluate the selection qualitatively.
Figure~\ref{fig:semantic_selection} shows the instances selected by our algorithm from the first frame of a set of sequences.
(a) - (c) show cases where the selection is correct and accurate, while in (d) - (e) the semantics is correct
but the mask is not accurate.
(f) - (h) are cases where the correct category is not present in the vocabulary of PASCAL, but there is a reasonable match that provides us with a good mask (camel vs horse, goat vs dog, chair vs homemade trolley).
Finally, (i) and (j) show cases where the algorithm selects two instances where there should only be one, in the latter possibly due to the fact that the dancer is upside-down and so there is no correct instance segmentation in the 
proposals.

\begin{table}
\setlength{\tabcolsep}{2pt} 
\center
\footnotesize
\rowcolors{4}{rowblue}{white}
\resizebox{\linewidth}{!}{%
\sisetup{detect-all=true}
\begin{tabular}{llS[table-format=2.1]S[table-format=2.1]S[table-format=2.1]S[table-format=2.1]S[table-format=2.1]}
\toprule
                            & & \multicolumn{2}{c}{Semantic} &    \si{Concatenate}              & \si{Conditional} &\\
\multicolumn{2}{c}{Measure} & \si{Automatic} & \si{Oracle} & \si{Features} & \si{Classifier} & Ours\\
\cmidrule(lr){1-2} \cmidrule(lr){3-4} \cmidrule(lr){5-6}\cmidrule(lr){7-7}
                                 & Mean $\mathcal{M} \uparrow$     &          68.9 &          70.1 &          76.7 &          82.3 &\bfseries 84.0 \\
\cellcolor{rowblue}$\mathcal{J}$ & Recall $\mathcal{O} \uparrow$   &          85.5 &          87.8 &          91.0 &          95.3 &\bfseries 96.1 \\
                                 & Decay $\mathcal{D} \downarrow$  &\bfseries 3.3 &          3.5 &          19.4 &          8.7 &          4.9 \\
\hline
                                 & Mean $\mathcal{M} \uparrow$     &          58.5 &          59.5 &          76.9 &          82.3 &\bfseries 83.0 \\
\cellcolor{white}$\mathcal{F}$ & Recall $\mathcal{O} \uparrow$   &          63.0 &          66.2 &          90.4 &\bfseries 94.3 &          93.8 \\
                                 & Decay $\mathcal{D} \downarrow$  &\bfseries 3.0 &          4.2 &          22.4 &          12.7 &          7.7 \\
\hline
\cellcolor{rowblue}$\mathcal{T}$ & Mean $\mathcal{M} \downarrow$   &          30.5 &          33.0 &          34.8 &\bfseries 23.9 &          25.6 \\
\bottomrule
\end{tabular}
}
\vspace{2mm}
\caption{\label{tab:ablation}
\textbf{Ablation results}: Comparison between the different baseline versions of our technique.}
\vspace{0mm}
\end{table}

\begin{table*}
\setlength{\tabcolsep}{4pt} 
\center
\footnotesize
\rowcolors{1}{white}{rowblue}
\resizebox{0.82\linewidth}{!}{%
\sisetup{detect-all=true}
\begin{tabular}{lS[table-format=2.1]@{\hspace{1.5mm}}>{\fontsize{6}{6}}S[table-format=2.1]S[table-format=2.1]@{\hspace{1.5mm}}>{\fontsize{6}{6}}S[table-format=2.1]S[table-format=2.1]@{\hspace{1.5mm}}>{\fontsize{6}{6}}S[table-format=2.1]S[table-format=2.1]@{\hspace{1.5mm}}>{\fontsize{6}{6}}S[table-format=2.1]S[table-format=2.1]@{\hspace{1.5mm}}>{\fontsize{6}{6}}S[table-format=2.1]S[table-format=2.1]@{\hspace{1.5mm}}>{\fontsize{6}{6}}S[table-format=2.1]S[table-format=2.1]@{\hspace{1.5mm}}>{\fontsize{6}{6}}S[table-format=2.1]S[table-format=2.1]@{\hspace{1.5mm}}>{\fontsize{6}{6}}S[table-format=2.1]S[table-format=2.1]@{\hspace{1.5mm}}>{\fontsize{6}{6}}S[table-format=2.1]S[table-format=2.1]@{\hspace{1.5mm}}>{\fontsize{6}{6}}S[table-format=2.1]}
\toprule
Attr& \multicolumn{2}{c}{Ours} & \multicolumn{2}{c}{\nosvos} & \multicolumn{2}{c}{\nmsk} & \multicolumn{2}{c}{\nofl} & \multicolumn{2}{c}{\nbvs} & \multicolumn{2}{c}{\nfcp} & \multicolumn{2}{c}{\njmp} & \multicolumn{2}{c}{\nhvs} & \multicolumn{2}{c}{\nsea} & \multicolumn{2}{c}{\ntsp} \\
\midrule
LR	&\bfseries 82.7 &\itshape 1.6	&       77.2 &\itshape 3.5	&       76.3 &\itshape 4.6	&       45.9 &\itshape 29.5	&       43.0 &\itshape 22.6	&       48.4 &\itshape 13.4	&       41.5 &\itshape 20.6	&       34.9 &\itshape 26.4	&       32.8 &\itshape 23.5	&       20.3 &\itshape 15.5	\\
SV	&\bfseries 81.3 &\itshape 4.4	&       74.3 &\itshape 9.1	&       73.6 &\itshape 10.2	&       53.2 &\itshape 24.7	&       43.3 &\itshape 27.8	&       46.7 &\itshape 19.5	&       48.8 &\itshape 13.7	&       35.7 &\itshape 31.6	&       41.4 &\itshape 15.0	&       18.7 &\itshape 22.1	\\
SC	&\bfseries 76.4 &\itshape 11.6	&       70.6 &\itshape 14.2	&       70.2 &\itshape 14.6	&       57.9 &\itshape 15.6	&       55.1 &\itshape 7.5	&       48.1 &\itshape 15.8	&       46.1 &\itshape 16.7	&       40.8 &\itshape 21.3	&       36.3 &\itshape 21.7	&       18.4 &\itshape 20.8	\\
FM	&\bfseries 80.0 &\itshape 6.1	&       76.5 &\itshape 5.1	&       74.8 &\itshape 7.6	&       49.6 &\itshape 28.2	&       44.8 &\itshape 23.3	&       50.7 &\itshape 11.9	&       45.2 &\itshape 18.0	&       34.5 &\itshape 31.0	&       30.9 &\itshape 30.1	&       12.5 &\itshape 29.9	\\
DB	&       72.6 &\itshape 13.3	&\bfseries 74.3 &\itshape 6.5	&       74.1 &\itshape 6.6	&       44.3 &\itshape 27.9	&       31.9 &\itshape 33.0	&       53.4 &\itshape 5.9	&       40.7 &\itshape 19.1	&       42.9 &\itshape 13.9	&       31.1 &\itshape 22.7	&       14.7 &\itshape 20.3	\\
MB	&\bfseries 79.1 &\itshape 8.9	&       73.7 &\itshape 11.0	&       73.4 &\itshape 11.4	&       55.5 &\itshape 22.8	&       53.7 &\itshape 11.5	&       50.9 &\itshape 13.6	&       50.9 &\itshape 11.1	&       42.3 &\itshape 22.5	&       39.3 &\itshape 20.3	&       17.4 &\itshape 26.4	\\
OCC	&\bfseries 81.3 &\itshape 3.7	&       77.2 &\itshape 3.7	&       75.5 &\itshape 6.0	&       67.3 &\itshape 1.0	&       67.3 &\itshape -10.4	&       49.2 &\itshape 13.2	&       45.1 &\itshape 16.9	&       48.7 &\itshape 8.5	&       38.2 &\itshape 17.5	&       23.9 &\itshape 11.5	\\
AC	&\bfseries 83.6 &\itshape 0.6	&       80.6 &\itshape -1.2	&       79.8 &\itshape -0.1	&       56.6 &\itshape 17.6	&       48.6 &\itshape 17.6	&       52.8 &\itshape 8.6	&       52.4 &\itshape 7.0	&       41.4 &\itshape 20.4	&       43.2 &\itshape 11.1	&       19.2 &\itshape 19.5	\\
\bottomrule
\end{tabular}}
\vspace{2mm}
\caption{\label{tab:evaltableattr} \textbf{Attribute-based performance}: Impact of the attributes of the sequences on the results.
For each attribute, results on the sequences with that particular feature and the gain with respect to those on the set of sequences without the attribute. LR stands for low resolution, SV for scale variation, SC for shape complexity, FM for fast motion, DB for dynamic background, MB for motion blur, OCC for occlusions, and AC for appearance change.
}
\vspace{-3mm}
\end{table*}

The first two columns of Table~\ref{tab:ablation} evaluate the second step of our algorithm: the semantic propagation.
The instances propagated in all the frames according to the semantic selection are evaluated directly as a final result without post-processing.
The first column comes from the instances selected by our algorithm while the second one from an oracle that has access to the ground truth of the whole sequence and selects the instances as in the first frame.
The small loss in performance of our algorithm with respect to the oracle shows that our semantic propagation algorithm makes the best of the semantic instance segmentation algorithm.

\paragraph*{Ablation Analysis:}
This section compares different ablated versions of our algorithm to measure how much each component adds. As a common practice, post-processing can be added to the raw output from CNN to improve the result. In this work, we use bilateral filter to smooth the output. 
The right-most column of Table~\ref{tab:ablation} shows our final result using the conditional classifier and  post-processing.
Comparing to the second column to the right (Conditional classifier), the post-processing adds a final 1.7 points to the results. 
On its left, we show the result obtained by concatenating the instance masks as a new feature map before the final classifier. We see a drop of performance (-5.6) that justifies our conditional classifier architecture.
To measure how much the refinement step adds, the two left-most columns evaluate the semantic instances directly, either selected by our method or by an oracle.
We observe a drop of -12.2 points even when using the oracle, which shows the inaccuracy of the instance segmentations and justifies the refinement step.

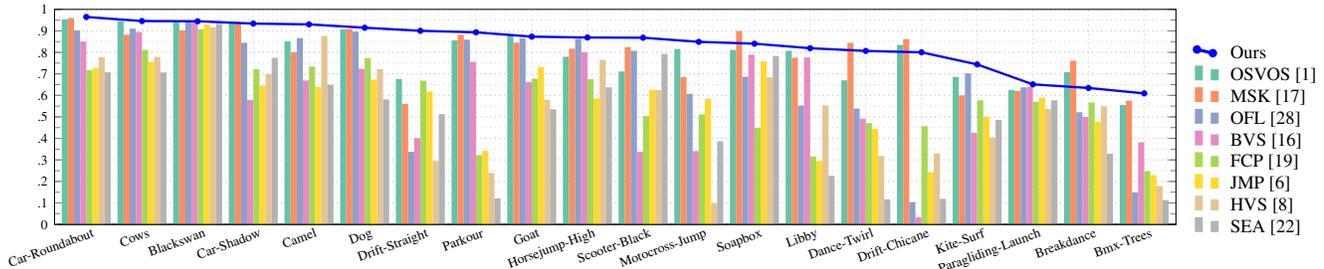
\begin{figure*}
\pgfplotstableread{data/per_seq_mean_J.txt}\perseqdata
\mbox{%
\begin{minipage}{0.90\textwidth}
  \resizebox{\textwidth}{!}{%
    \begin{tikzpicture}
        \begin{axis}[set layers,width=1.4\textwidth,height=0.35\textwidth,
                ybar=0pt,bar width=0.11,
                grid=both,
				grid style=dotted,
                minor ytick={0,0.05,...,1.1},
    			ytick={0,0.1,...,1.1},
			    yticklabels={0,.1,.2,.3,.4,.5,.6,.7,.8,.9,1},
                ymin=0, ymax=1,
                xtick = data, x tick label style={rotate=20,anchor=north east,xshift=7pt,yshift=5pt},
                xticklabels from table={\perseqdata}{Seq},
                major x tick style = transparent,
                enlarge x limits=0.03,
                font=\scriptsize,
                ]
             \addplot[draw opacity=0,fill=Set2-8-1,mark=none,legend image post style={yshift=-0.1em}] table[x expr=\coordindex,y=OSVOS]{\perseqdata};
             \label{fig:perseq:osvos}
             \addplot[draw opacity=0,fill=Set2-8-2,mark=none,legend image post style={yshift=-0.1em}] table[x expr=\coordindex,y=MSK]{\perseqdata};
             \label{fig:perseq:msk}
            \addplot[draw opacity=0,fill=Set2-8-3,mark=none,legend image post style={yshift=-0.1em}] table[x expr=\coordindex,y=OFL]{\perseqdata};
            \label{fig:perseq:ofl}
            \addplot[draw opacity=0,fill=Set2-8-4,mark=none,legend image post style={yshift=-0.1em}] table[x expr=\coordindex,y=BVS]{\perseqdata};
            \label{fig:perseq:bvs}
            \addplot[draw opacity=0,fill=Set2-8-5,mark=none,legend image post style={yshift=-0.1em}] table[x expr=\coordindex,y=FCP]{\perseqdata};
            \label{fig:perseq:fcp}
            \addplot[draw opacity=0,fill=Set2-8-6,mark=none,legend image post style={yshift=-0.1em}] table[x expr=\coordindex,y=JMP]{\perseqdata};
            \label{fig:perseq:jmp}
            \addplot[draw opacity=0,fill=Set2-8-7,mark=none,legend image post style={yshift=-0.1em}] table[x expr=\coordindex,y=HVS]{\perseqdata};
            \label{fig:perseq:hvs}
            \addplot[draw opacity=0,fill=Set2-8-8,mark=none,legend image post style={yshift=-0.1em}] table[x expr=\coordindex,y=SEA]{\perseqdata};
			\label{fig:perseq:sea}
            \addplot[blue,sharp plot,update limits=false,mark=*,mark size=1,line width=1.2pt, legend image post style={yshift=-0.4em}] table[x expr=\coordindex,y=SGV]{\perseqdata};
            \label{fig:perseq:sgv}
            
            \addplot[black,sharp plot,update limits=false] coordinates{(-0.5,0) (20.5,0)};
        \end{axis}
    \end{tikzpicture}
    }
    \end{minipage}
    \hspace{0mm}
    \begin{minipage}{0.07\textwidth}
    \scriptsize
    \begin{tabular}{@{}l@{\hspace{1.5mm}}l}
    \ref{fig:perseq:sgv}&Ours\\
    \ref{fig:perseq:osvos}&\nosvos~\cite{\osvos}\\
    \ref{fig:perseq:msk}&\nmsk~\cite{\msk}\\
    \ref{fig:perseq:ofl} &\nofl~\cite{\ofl}\\
    \ref{fig:perseq:bvs} &\nbvs~\cite{\bvs}\\
    \ref{fig:perseq:fcp} &\nfcp~\cite{\fcp}\\
    \ref{fig:perseq:jmp} &\njmp~\cite{\jmp}\\
    \ref{fig:perseq:hvs} &\nhvs~\cite{\hvs}\\
    \ref{fig:perseq:sea} &\nsea~\cite{\sea}
    \end{tabular}
    \end{minipage}
    }
    \vspace{-1mm}
    \caption{\label{fig:perseq}\textbf{Per-sequence evaluation}: Per-sequence results of region similarity ($\J$) on the DAVIS validation set, sorted by \ours' results.}
\end{figure*}

\paragraph*{Comparison to the State of the Art:}

Table~\ref{tab:soa} shows the comparison of our technique with the state of the art in the validation set of DAVIS.
We are around 4.2 points above \nosvos{} and \nmsk{} (the latest and best-performing techniques) in terms of mean $\J$ values.
Most notably, the decay values of \ours{} are significantly better, which means that the quality of our results does not decrease significantly as the sequence evolves. 
We attribute this behavior to the incorporation of the semantic prior, which is invariant through the sequence and helps keeping the results accurate.
It is also significant that we decrease the temporal instability ($\T$) noticeably from \nosvos{} and get closer to \nmsk{}, which propagates masks from frame to frame and uses optical flow so it is by design more stable. Instead, we process each frame independently.

Table~\ref{tab:evaltableattr} shows the per-attribute comparison in DAVIS, that is, the mean results on a subset of sequences where a certain challenging attribute is present.
\ours{} is the best performing in all attributes except dynamic background (by a close margin).
The increase of performance when each attribute is not present (small numbers in italics) is significantly low, which shows that \ours{} is the most robust method to the different challenges.

Figure~\ref{fig:perseq} breaks down the performance for each of the sequences in the validation set of DAVIS.
In columns, all state-of-the-art techniques are shown, and our result is overlaid with a line.
\ours{} is the best performing in the majority of the sequences, and only in four sequences the results go below 80\% accuracy, being 60\% the worse result.
Especially noticeable are the improvements in drift-straight or motocross-jump, where the objects change from frontal views to a lateral views (see Figure~\ref{fig:example}), which confuses the appearance models.
In the camel sequence we also improve significantly, in this case because a second camel appears from behind the first with the same appearance. We are able to discard it because we enforce the semantic prior that only one instance has to be selected (as in the given first frame).

Finally, the performance in the Youtube-objects dataset is evaluated as suggested in the original paper: the frames with empty masks are discarded and the performance is averaged over categories.
Table~\ref{tab:youtube} shows the evaluation per-category and the global mean.
\ours{} obtains the best results in 6 out of 10 categories, and is 2.5 points better than the second-performing technique. 

\begin{table}
\centering
\rowcolors{2}{white}{rowblue}
\resizebox{1\linewidth}{!}{%
\begin{tabular}{lccccccc}
\toprule
Categ.    & Ours      & \nosvos       & \nmsk    & \nofl      & \njfs      & \nbvs      & \nafs\\
\midrule
Aeroplane &    \ 89.5 &    \ 88.2 &    \ 86.0 & \bf\ 89.9 &    \ 89.0 &    \ 86.8 &    \ 79.9\\
Bird      & \bf\ 86.5 &    \ 85.7 &    \ 85.6 &    \ 84.2 &    \ 81.6 &    \ 80.9 &    \ 78.4\\
Boat      & \bf\ 80.9 &    \ 77.5 &    \ 78.8 &    \ 74.0 &    \ 74.2 &    \ 65.1 &    \ 60.1\\
Car       & \bf\ 84.3 &    \ 79.6 &    \ 78.8 &    \ 80.9 &    \ 70.9 &    \ 68.7 &    \ 64.4\\
Cat       & \bf\ 77.2 &    \ 70.8 &    \ 70.1 &    \ 68.3 &    \ 67.7 &    \ 55.9 &    \ 50.4\\
Cow       &    \ 78.6 &    \ 77.8 &    \ 77.7 & \bf\ 79.8 &    \ 79.1 &    \ 69.9 &    \ 65.7\\
Dog       & \bf\ 82.5 &    \ 81.3 &    \ 79.2 &    \ 76.6 &    \ 70.3 &    \ 68.5 &    \ 54.2\\
Horse     & \bf\ 73.2 &    \ 72.8 &    \ 71.7 &    \ 72.6 &    \ 67.8 &    \ 58.9 &    \ 50.8\\
Motorbike &    \ 72.7 &    \ 73.5 &    \ 65.6 & \bf\ 73.7 &    \ 61.5 &    \ 60.5 &    \ 58.3\\
Train     &    \ 82.9 &    \ 75.7 & \bf\ 83.5 &    \ 76.3 &    \ 78.2 &    \ 65.2 &    \ 62.4\\
\midrule
Mean      & \bf\ 80.8 &    \ 78.3 &    \ 77.7 &    \ 77.6 &    \ 74.0 &    \ 68.0 &    \ 62.5\\
\bottomrule
\end{tabular}}
\vspace{2mm}
\caption{\textbf{Youtube-Objects evaluation}: Per-category mean intersection over union ($\J$).}
\label{tab:youtube}
\end{table}

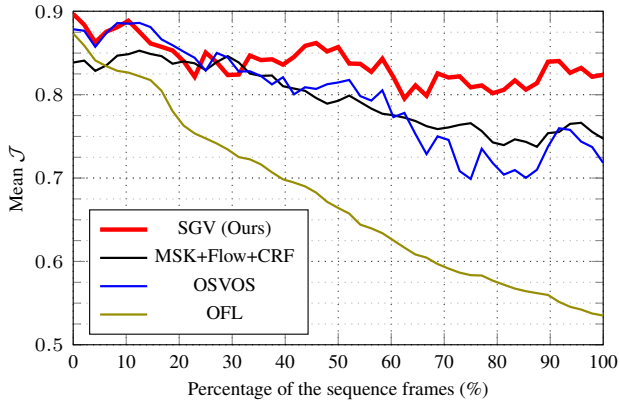
\begin{figure}
\resizebox{\linewidth}{!}{%
\begin{tikzpicture}[/pgfplots/width=1\linewidth, /pgfplots/height=0.7\linewidth]
    \begin{axis}[ymin=0.5,ymax=0.9,xmin=0,xmax=1,enlargelimits=false,
        xlabel=Percentage of the sequence frames (\%),
        ylabel=Mean $\J$,
        font=\scriptsize, grid=both,
        legend pos= south west,
        grid style=dotted,
        axis equal image=false,
        ytick={0.1,0.2,...,0.9},
        xtick={0,0.1,...,1.1},
        xticklabels={0,10,20,30,40,50,60,70,80,90,100},
        minor ytick={0.1,0.125,...,0.9},
        major grid style={white!20!black},
        minor grid style={white!70!black},
        xlabel shift={-2pt},
        ylabel shift={-3pt}]
                 \addplot+[red,solid,mark=none, ultra thick] table[x=time,y=OURS] {data/performance_drop_curve.txt};
                 \addlegendentry{SGV (Ours)}
         \addplot+[black,solid,mark=none, thick] table[x=time,y=Masktrack] {data/performance_drop_curve.txt};
                  \addlegendentry{MSK+Flow+CRF}
         \addplot+[blue,solid,mark=none, thick] table[x=time,y=OSVOS] {data/performance_drop_curve.txt};
         \addlegendentry{OSVOS}
         \addplot+[olive,solid,mark=none, thick] table[x=time,y=OFL] {data/performance_drop_curve.txt};
         \addlegendentry{OFL}

	 \end{axis}
   \end{tikzpicture}}
   \caption{\textbf{Decay of the quality with time}: Performance of various methods with respect to the \textit{time} axis.}
   \label{fig:decay}
\end{figure}

\paragraph*{\textbf{Performance Decay}:}
As indicated by the $\J$-Decay and $\F$-Decay values in Table~\ref{tab:soa}, \ours{} exhibits a significant better ability to maintain performance as frames evolve, and we interpret that this is so thanks to the injected semantic prior.
To further highlight this result and analyze it more in detail, 
Figure~\ref{fig:decay} shows the evolution of $\J$ as the sequence advances, to examine how the performance drops over time.
Since the videos in DAVIS are of different length, we normalize them to $[0,100]$ as a percentage of the sequence length.
We then compute the mean $\J$ curve among all video sequences.
As can be seen from Figure~\ref{fig:decay}, our method is significantly more stable in terms of performance drop compared to the competing methods.
More specifically, our method is able to maintain a good performance of $82.38$ even at the end of the video, while the competing methods are at $74.72$, $53.50$ and $71.80$, for MSK+Flow+CRF, OFL and OSVOS, respectively.
We therefore achieve an $8\%$ improvement compared to the closest competitor.

We also report the lowest point of the curve which indicates the worst performance across the video.
Based on this metrics, \ours{} is at $79.10$, while for semantic-blind methods, the numbers are $73.62$, $53.50$ and $69.55$.

The results therefore confirm that the semantic prior we introduce can mitigate the performance drop caused by appearance change, while maintaining high fidelity in details.
The semantic information is particularly helpful in the later stage of videos where more dramatic appearance changes are present.
It is also noticeable that there are more oscillations in the curves of \nosvos{} and \ours{}.
This is probably due to the fact that they do not make use of any temporal information, while the other techniques do.

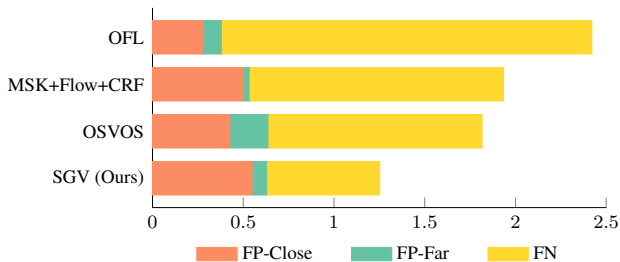
\begin{figure}[b]
\centering
\resizebox{\linewidth}{!}{\begin{tikzpicture}
\begin{axis}[
    xbar stacked,
    legend style={
    legend columns=3,
        at={(xticklabel cs:0.5)},
        anchor=north,
        draw=none
    },
    ytick=data,
    axis y line*=none,
    axis x line*=bottom,
    tick label style={font=\footnotesize},
    legend style={font=\footnotesize},
    label style={font=\footnotesize},
	legend style={/tikz/every even column/.append style={column sep=5mm}},
    width=\linewidth,
    bar width=5mm,
    yticklabels={SGV (Ours), OSVOS, MSK+Flow+CRF, OFL},
    xmin=0,
    xmax=2.5,
    area legend,
    y=7mm,
    enlarge y limits={abs=0.625}]
\addplot[Set2-8-2,fill=Set2-8-2] coordinates
{(0.5488,0) (0.4264,1) (0.4994,2) (0.277,3)};
\addplot[Set2-8-1,fill=Set2-8-1] coordinates
{(0.08039,0) (0.2097,1) (0.03317,2) (0.102,3)};
\addplot[Set2-8-6,fill=Set2-8-6] coordinates
{(0.6227,0) (1.18,1) (1.402,2) (2.042,3)};
\legend{FP-Close, FP-Far, FN}
\end{axis}  
\end{tikzpicture}}
\vspace{-1mm}
\caption{\textbf{Error analysis of our method}: Errors divided into False Positives (FP-Close and FP-Far) and False Negatives (FN). Values are percentage (\%) of FP-Close, FP-Far or FN pixels in a sequence.}
\label{fig:error_stats}
\end{figure}

\paragraph*{\textbf{Misclassified-Pixels Analysis}:}
Figure~\ref{fig:error_stats} shows the error analysis of our method.
We divide the misclassified pixels in three categories:
Close False Positives (FP-Close), Far False Positives (FP-Far) and False Negatives (FN):
(i) FP-Close are those near the contour of the object of interest, so contour inaccuracies,
(ii) FP-Far reveal if the method detects other objects or blobs apart from the object of interest, and
(iii) FN tell us if we miss a part of the object during the sequence.
The measure in the plot is the percentage of pixels in a sequence that fall in a certain category.

\begin{figure*}
\centering
\resizebox{\textwidth}{!}{%
	  \setlength{\fboxsep}{0pt}
      \rotatebox{90}{\hspace{4.5mm}Drift-Chicane\vphantom{p}}
      \fbox{\includegraphics[width=0.3\textwidth]{ims/examples/drift-straight/00000.png}}   
      \fbox{\includegraphics[width=0.3\textwidth]{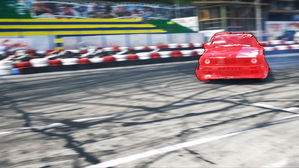}}
      \fbox{\includegraphics[width=0.3\textwidth]{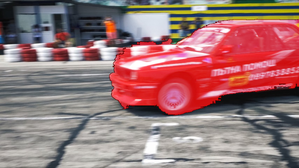}}
      \fbox{\includegraphics[width=0.3\textwidth]{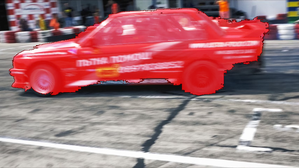}}
      \fbox{\includegraphics[width=0.3\textwidth]{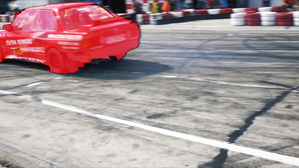}}
      }\\[1mm]
\resizebox{\textwidth}{!}{%
	  \setlength{\fboxsep}{0pt}
      \rotatebox{90}{\hspace{2.5mm}Motocross-Jump}
      \fbox{\includegraphics[width=0.3\textwidth]{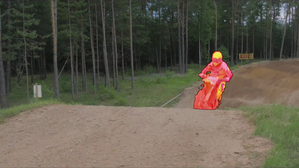}}
      \fbox{\includegraphics[width=0.3\textwidth]{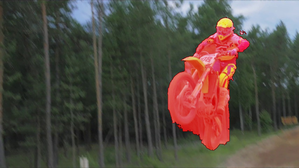}}
      \fbox{\includegraphics[width=0.3\textwidth]{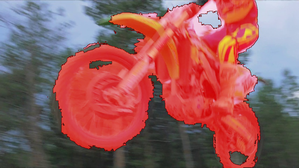}}
      \fbox{\includegraphics[width=0.3\textwidth]{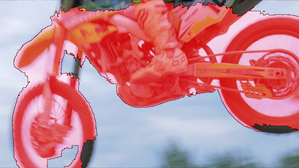}}
      \fbox{\includegraphics[width=0.3\textwidth]{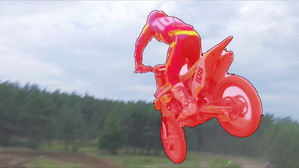}}
      }\\[1mm]
\resizebox{\textwidth}{!}{%
	  \setlength{\fboxsep}{0pt}
      \rotatebox{90}{\hspace{7.5mm}Kite-Surf\vphantom{p}}
      \fbox{\includegraphics[width=0.3\textwidth]{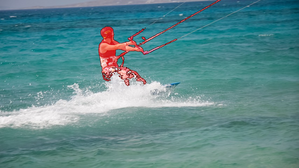}}
      \fbox{\includegraphics[width=0.3\textwidth]{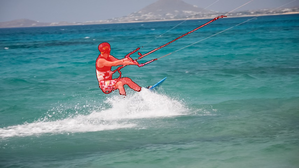}}
      \fbox{\includegraphics[width=0.3\textwidth]{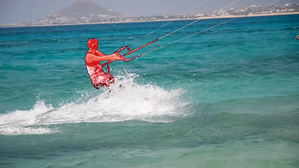}}
      \fbox{\includegraphics[width=0.3\textwidth]{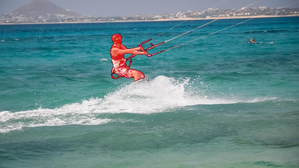}}
      \fbox{\includegraphics[width=0.3\textwidth]{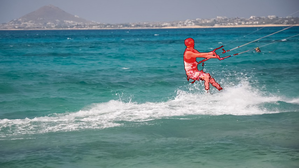}}
      }\\[1mm]
\resizebox{\textwidth}{!}{%
	  \setlength{\fboxsep}{0pt}
      \rotatebox{90}{\hspace{8.5mm}Parkour\vphantom{p}}
      \fbox{\includegraphics[width=0.3\textwidth]{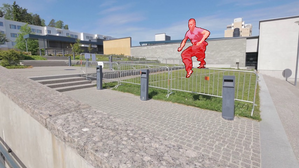}}
      \fbox{\includegraphics[width=0.3\textwidth]{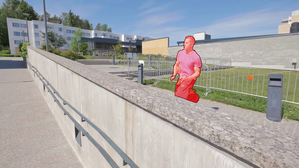}}
      \fbox{\includegraphics[width=0.3\textwidth]{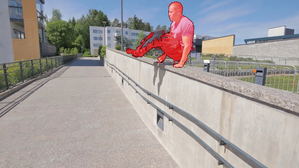}}
      \fbox{\includegraphics[width=0.3\textwidth]{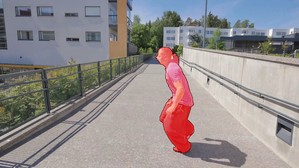}}
      \fbox{\includegraphics[width=0.3\textwidth]{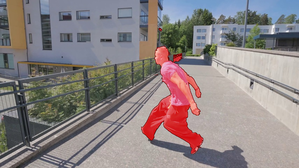}}
      }\\[1mm]
\resizebox{\textwidth}{!}{%
	  \setlength{\fboxsep}{0pt}
      \rotatebox{90}{\hspace{1.5mm}Car-Roundabout\vphantom{p}}
      \fbox{\includegraphics[width=0.3\textwidth]{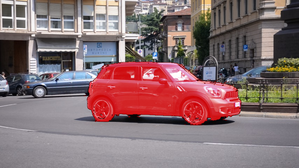}}
      \fbox{\includegraphics[width=0.3\textwidth]{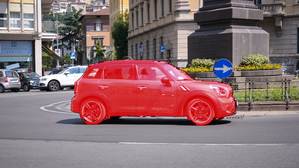}}
      \fbox{\includegraphics[width=0.3\textwidth]{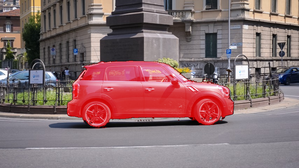}}
      \fbox{\includegraphics[width=0.3\textwidth]{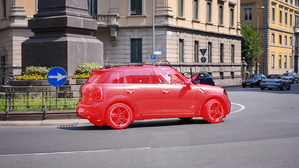}}
      \fbox{\includegraphics[width=0.3\textwidth]{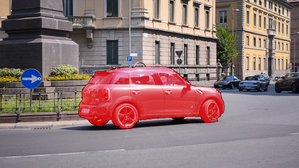}}
      }\\[1mm]
\resizebox{\textwidth}{!}{%
	  \setlength{\fboxsep}{0pt}
      \rotatebox{90}{\hspace{8.5mm}Camel\vphantom{p}}
      \fbox{\includegraphics[width=0.3\textwidth]{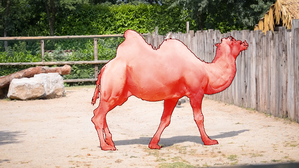}}
      \fbox{\includegraphics[width=0.3\textwidth]{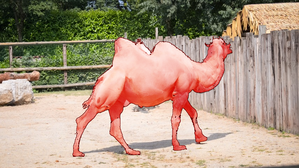}}
      \fbox{\includegraphics[width=0.3\textwidth]{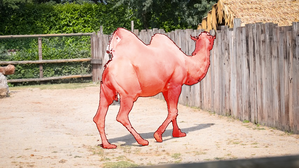}}
      \fbox{\includegraphics[width=0.3\textwidth]{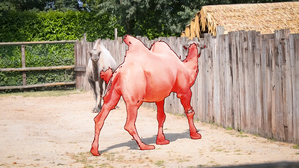}}
      \fbox{\includegraphics[width=0.3\textwidth]{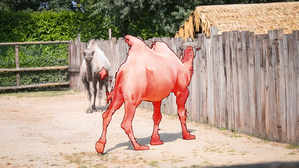}}
      }\\[1mm]
\resizebox{\textwidth}{!}{%
	  \setlength{\fboxsep}{0pt}
      \rotatebox{90}{\hspace{2.5mm}Horsejump-High\vphantom{p}}
      \fbox{\includegraphics[width=0.3\textwidth]{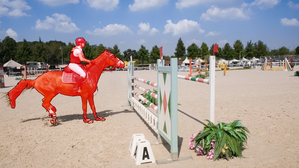}}
      \fbox{\includegraphics[width=0.3\textwidth]{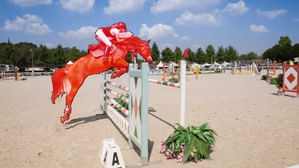}}
      \fbox{\includegraphics[width=0.3\textwidth]{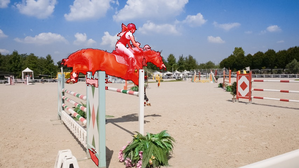}}
      \fbox{\includegraphics[width=0.3\textwidth]{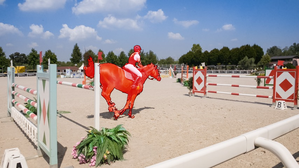}}
      \fbox{\includegraphics[width=0.3\textwidth]{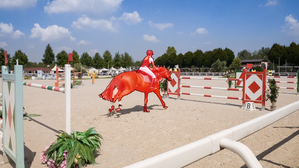}}
      }\\[1mm]   
\resizebox{\textwidth}{!}{%
	  \setlength{\fboxsep}{0pt}
      \rotatebox{90}{\hspace{6.5mm}Bmx-Trees\vphantom{p}}
      \fbox{\includegraphics[width=0.3\textwidth]{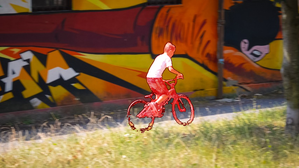}}
      \fbox{\includegraphics[width=0.3\textwidth]{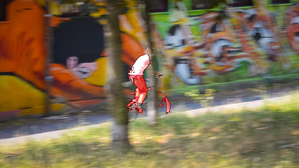}}
      \fbox{\includegraphics[width=0.3\textwidth]{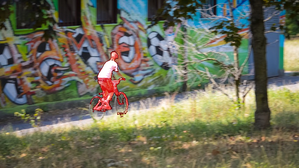}}
      \fbox{\includegraphics[width=0.3\textwidth]{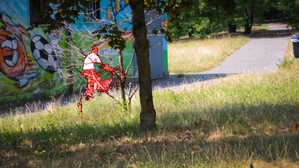}}
      \fbox{\includegraphics[width=0.3\textwidth]{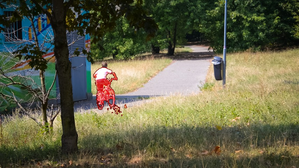}}
      }
\caption{\textbf{Qualitative results}: \ours{} results on a variety of representatives sequences.}
\label{fig:qualitative}
\vspace{-2mm}
\end{figure*}

\ours{}'s main strength, compared to the other state of the art methods, is reducing the number of false negatives considerably.
We believe this is due to \ours{}'s ability to \textit{complete} the object of interest when parts that were occluded in the first frame become visible, thanks to the semantic concept of instance.
On the other hand, the output of the instance segmentation network that we are currently using, MNC, is very imprecise on the boundaries of the objects, and even though our conditional classifier is able to recover in part, FP-Close is slightly worse than that of the competition.
On the plus side, since the instance segmentation is an independent input to our algorithm, we will probably directly benefit from better instance segmentation algorithms.

\paragraph*{Timing:}
Our method consists of an instance segmentation CNN as well as an appearance-model CNN.
These two can be run in parallel, as just the final conditional classifier combines the output of the two networks.
Using an NVidia Titan X GPU, the appearance model takes 120\,ms and the instance segmentation CNN (MNC) takes 360\,ms.
After that, the fast bilateral solver to refine the results takes 187\,ms on a CPU.
Overall, the time per frame is 547\,ms.
We would like to point out that our model enables to plug and play another instance segmentation method and therefore the timing could be improved when a faster method is available.

\paragraph*{Qualitative Results:} Figure~\ref{fig:qualitative} shows some qualitative results of \ours{}.
The first row shows a sequence with significant change of appearance due to different points of view, where we correctly segment front, side, and back views of the car thanks to the semantic prior.

\section{Conclusions}

This paper presents \ours{}, a semi-supervised video object segmentation technique that leverages an instance segmentation algorithm to guide the appearance model computed on the first frame.
Given the initial mask, the semantics of the object is estimated by overlapping the proposals of segmented instances in the scene.
This information is propagated throughout the sequence and the most promising instances that match with the predicted semantics at each frame are taken as the semantic prior for the segmentation.
The appearance model is then combined with the semantic prior by means of a conditional classifier as a trainable module in a CNN.
\ours{} shows state-of-the-art results in DAVIS and Youtube-Objects and runs in half a second per frame.

\section*{Acknowledgments}
This work was supported by the Swiss Commission for Technology and Innovation (CTI, Grant No. 19015.1 PFES-ES, NeGeVA). The authors gratefully acknowledge support by armasuisse, and thank NVIDIA Corporation for donating the GPUs used in this project.

{\small
\bibliographystyle{ieee}
\bibliography{ICCV2017}

\begin{thebibliography}{10}\itemsep=-1pt

\bibitem{Caelles2017}
S.~Caelles, K.-K. Maninis, J.~Pont-Tuset, L.~Leal-Taix\'e, D.~Cremers, and
  L.~{Van Gool}.
\newblock One-shot video object segmentation.
\newblock In {\em CVPR}, 2017.

\bibitem{Chang2013}
J.~Chang, D.~Wei, and J.~W. {Fisher III}.
\newblock A video representation using temporal superpixels.
\newblock In {\em CVPR}, 2013.

\bibitem{dai2015convolutional}
J.~Dai, K.~He, and J.~Sun.
\newblock Convolutional feature masking for joint object and stuff
  segmentation.
\newblock In {\em CVPR}, 2015.

\bibitem{dai2016instance}
J.~Dai, K.~He, and J.~Sun.
\newblock Instance-aware semantic segmentation via multi-task network cascades.
\newblock In {\em CVPR}, 2016.

\bibitem{dantone2012real}
M.~Dantone, J.~Gall, G.~Fanelli, and L.~Van~Gool.
\newblock Real-time facial feature detection using conditional regression
  forests.
\newblock In {\em CVPR}, 2012.

\bibitem{Fan2015}
Q.~Fan, F.~Zhong, D.~Lischinski, D.~Cohen-Or, and B.~Chen.
\newblock Jumpcut: Non-successive mask transfer and interpolation for video
  cutout.
\newblock {\em {ACM} Trans. Graph.}, 34(6), 2015.

\bibitem{girshick2014rich}
R.~Girshick, J.~Donahue, T.~Darrell, and J.~Malik.
\newblock Rich feature hierarchies for accurate object detection and semantic
  segmentation.
\newblock In {\em CVPR}, 2014.

\bibitem{Grundmann2010}
M.~Grundmann, V.~Kwatra, M.~Han, and I.~A. Essa.
\newblock Efficient hierarchical graph-based video segmentation.
\newblock In {\em CVPR}, 2010.

\bibitem{hane2013joint}
C.~Hane, C.~Zach, A.~Cohen, R.~Angst, and M.~Pollefeys.
\newblock Joint 3d scene reconstruction and class segmentation.
\newblock In {\em Proceedings of the IEEE Conference on Computer Vision and
  Pattern Recognition}, pages 97--104, 2013.

\bibitem{hariharan2014simultaneous}
B.~Hariharan, P.~Arbel{\'a}ez, R.~Girshick, and J.~Malik.
\newblock Simultaneous detection and segmentation.
\newblock In {\em ECCV}, 2014.

\bibitem{Jain2014}
S.~D. Jain and K.~Grauman.
\newblock Supervoxel-consistent foreground propagation in video.
\newblock In {\em ECCV}, 2014.

\bibitem{li2016iterative}
K.~Li, B.~Hariharan, and J.~Malik.
\newblock Iterative instance segmentation.
\newblock In {\em CVPR}, pages 3659--3667, 2016.

\bibitem{li2016fully}
Y.~Li, H.~Qi, J.~Dai, X.~Ji, and Y.~Wei.
\newblock Fully convolutional instance-aware semantic segmentation.
\newblock {\em arXiv preprint arXiv:1611.07709}, 2016.

\bibitem{liu2010single}
B.~Liu, S.~Gould, and D.~Koller.
\newblock Single image depth estimation from predicted semantic labels.
\newblock In {\em CVPR}, 2010.

\bibitem{Maninis2016}
K.~Maninis, J.~Pont-Tuset, P.~Arbel\'{a}ez, and L.~{Van Gool}.
\newblock Convolutional oriented boundaries.
\newblock In {\em ECCV}, 2016.

\bibitem{NicolasMaerki2016}
N.~Nicolas~M{\"a}rki, F.~Perazzi, O.~Wang, and A.~Sorkine-Hornung.
\newblock Bilateral space video segmentation.
\newblock In {\em CVPR}, 2016.

\bibitem{Perazzi2017}
F.~Perazzi, A.~Khoreva, R.~Benenson, B.~Schiele, and A.Sorkine-Hornung.
\newblock Learning video object segmentation from static images.
\newblock In {\em CVPR}, 2017.

\bibitem{Perazzi2016}
F.~Perazzi, J.~Pont-Tuset, B.~McWilliams, L.~{Van Gool}, M.~Gross, and
  A.~Sorkine-Hornung.
\newblock A benchmark dataset and evaluation methodology for video object
  segmentation.
\newblock In {\em CVPR}, 2016.

\bibitem{Perazzi2015}
F.~Perazzi, O.~Wang, M.~Gross, and A.~Sorkine-Hornung.
\newblock Fully connected object proposals for video segmentation.
\newblock In {\em ICCV}, 2015.

\bibitem{Pinheiro2016}
P.~O. Pinheiro, T.-Y. Lin, R.~Collobert, and P.~Doll{\'a}r.
\newblock Learning to refine object segments.
\newblock In {\em ECCV}, 2016.

\bibitem{Prest2012}
A.~Prest, C.~Leistner, J.~Civera, C.~Schmid, and V.~Ferrari.
\newblock Learning object class detectors from weakly annotated video.
\newblock In {\em CVPR}, 2012.

\bibitem{Ramakanth2014}
S.~A. Ramakanth and R.~V. Babu.
\newblock Seamseg: Video object segmentation using patch seams.
\newblock In {\em CVPR}, 2014.

\bibitem{romera2016recurrent}
B.~Romera-Paredes and P.~H.~S. Torr.
\newblock Recurrent instance segmentation.
\newblock In {\em ECCV}, 2016.

\bibitem{Russakovsky2015}
O.~Russakovsky, J.~Deng, H.~Su, J.~Krause, S.~Satheesh, S.~Ma, Z.~Huang,
  A.~Karpathy, A.~Khosla, M.~Bernstein, A.~C. Berg, and L.~Fei-Fei.
\newblock {ImageNet Large Scale Visual Recognition Challenge}.
\newblock {\em IJCV}, 2015.

\bibitem{ShankarNagaraja2015}
N.~Shankar~Nagaraja, F.~R. Schmidt, and T.~Brox.
\newblock Video segmentation with just a few strokes.
\newblock In {\em ICCV}, 2015.

\bibitem{SiZi15}
K.~Simonyan and A.~Zisserman.
\newblock Very deep convolutional networks for large-scale image recognition.
\newblock In {\em ICLR}, 2015.

\bibitem{sun2012conditional}
M.~Sun, P.~Kohli, and J.~Shotton.
\newblock Conditional regression forests for human pose estimation.
\newblock In {\em CVPR}, 2012.

\bibitem{Tsai2016}
Y.-H. Tsai, M.-H. Yang, and M.~J. Black.
\newblock Video segmentation via object flow.
\newblock In {\em CVPR}, 2016.

\bibitem{uijlings2015situational}
J.~R. Uijlings and V.~Ferrari.
\newblock Situational object boundary detection.
\newblock In {\em CVPR}, 2015.

\bibitem{Vijayanarasimhan2012}
S.~Vijayanarasimhan and K.~Grauman.
\newblock Active frame selection for label propagation in videos.
\newblock In {\em ECCV}, 2012.

\bibitem{XiTu15}
S.~Xie and Z.~Tu.
\newblock Holistically-nested edge detection.
\newblock In {\em ICCV}, 2015.

\bibitem{zagoruyko2016multipath}
S.~Zagoruyko, A.~Lerer, T.-Y. Lin, P.~O. Pinheiro, S.~Gross, S.~Chintala, and
  P.~Doll{\'a}r.
\newblock A multipath network for object detection.
\newblock {\em arXiv preprint arXiv:1604.02135}, 2016.

\end{thebibliography}
}

\end{document}